\documentclass[journal]{IEEEtran}
\usepackage{graphicx}
\usepackage{comment}
\usepackage{amsmath,amssymb} % define this before the line numbering.
\usepackage{color}
\usepackage{amsmath,amssymb,amsfonts}
\usepackage[ruled,vlined, linesnumbered]{algorithm2e}
\usepackage{cite}
\usepackage{textcomp}
\usepackage{xcolor}
\usepackage{multirow}
\usepackage{wrapfig,lipsum,booktabs}
\usepackage{makecell}
\usepackage{tabularx}

\usepackage[ruled,vlined]{algorithm2e}

\begin{document}
\title{QUANOS- Adversarial Noise Sensitivity Driven Hybrid Quantization of Neural Networks}

\author{Priyadarshini~Panda,~\IEEEmembership{Member,~IEEE},\\ Department of Electrical Engineering, Yale University, New Haven, CT, USA \\e-mail: priya.panda@yale.edu
       % <-this % stops a space
%\thanks{P. Panda is with the Department
%of Electrical Engineering, Yale University, New Haven,
%CT, 06510 USA e-mail: priya.panda@yale.edu.}% <-this % stops a space
\thanks{Accepted in ACM/IEEE International Symposium on Low Power Electronics and Design, 2020}% <-this % stops a space
}

\maketitle
\begin{abstract}
Deep Neural Networks (DNNs) have been shown to be vulnerable to adversarial attacks, wherein, a model gets fooled by applying slight perturbations on the input. With the advent of Internet-of-Things and the necessity to enable intelligence in embedded devices like mobile phones, smart watches etc., low-power and secure hardware implementation of DNNs is vital. In this paper, we investigate the use of quantization to potentially resist adversarial attacks. Several recent studies have reported remarkable results in reducing the energy requirement of a DNN through quantization. However, no prior work has considered the relationship between adversarial sensitivity of a DNN and its effect on quantization. We propose QUANOS- a framework that performs layer-specific hybrid quantization based on Adversarial Noise Sensitivity (ANS). We identify a novel noise stability metric (ANS) for DNNs, i.e., the sensitivity of each layer’s computation to adversarial noise. ANS allows for a principled way of determining optimal bit-width per layer that incurs adversarial robustness as well as energy-efficiency with minimal loss in accuracy. Essentially, QUANOS assigns layer significance based on its contribution to adversarial perturbation and accordingly scales the precision of the layers. A key advantage of QUANOS is that it does not rely on a pre-trained model and can be applied in the initial stages of training. We evaluate the benefits of QUANOS on precision scalable Multiply and Accumulate (MAC) hardware architectures with data gating and subword parallelism capabilities. Our experiments on CIFAR10, CIFAR100 datasets show that QUANOS outperforms homogenously quantized 8-bit precision baseline in terms of adversarial robustness ($3-4\%$ higher) while yielding improved compression ($>5\times$) and energy savings ($>2\times$) at iso-accuracy. At iso-compression rate, QUANOS yields significantly higher adversarial robustness ($>10\%$) than similar sized baseline against strong white-box attacks. We also find that combining QUANOS with state-of-the-art defense methods outperforms the state-of-the-art in robustness ($\sim5\%-16\%$ higher) against very strong attacks.
%For comparison, we use 16-bit and 8-bit precision models as baseline.
\end{abstract}


% \fi
\IEEEpeerreviewmaketitle

\section{Introduction}
Despite achieving super-human performance and classification accuracies on a variety of perception tasks including, vision, gaming among others \cite{lecun2015deep}, Deep Neural Networks (DNNs) have been shown to be adversarially vulnerable \cite{madry2017towards}. A DNN can be easily fooled into misclassifying an input with slight changes of pixel intensities. What is more worrying is that such slight, adversarial perturbations, while being imperceptible to human eyes mislead a DNN to misclassify with high confidence. This vulnerability severely limits the deployment and potential safe-use of DNNs for real world applications such as self-driving cars, malware detection, healthcare monitoring systems etc. \cite{carlini2019evaluating}. It is critical to ensure that the DNN models used in these applications are robust, as failure to do so can have disruptive consequences ranging from loss in revenue to loss of lives. Recent works \cite{panda2019discretization, lin2019defensive} presents quantization as a straightforward way of improving the intrinsic resilience of DNNs against adversarial attacks. Specifically, in \cite{panda2019discretization, lin2019defensive}, the authors showed that quantization, primarily used to reduce compute resource requirements of DNNs, also warrants security against certain level of adversarial perturbation, thereby, offering a key benefit of robustness in hardware implementation. Inspired by this, our work asks the following question: \textit{Can we perform layer-specific hybrid quantization of different layers of a DNN while optimally trading off between energy-accuracy-and-robustness?} 

Quantization is a popular compression technique used to reduce the number of bits required for encoding a DNN’s weights and activations. This in turn reduces the total computation energy as well as the data access energy in case of hardware implementation \cite{han2015deep, wu2016quantized}. The extreme case is that of a 1-bit quantized DNN. Such binary quantized models are usually trained from scratch to obtain competitive accuracy as that of a full-precision model \cite{rastegari2016xnor}. Recent works have also shown combining quantization with other techniques, such as pruning \cite{han2015deep}, yield higher compression rates which translate to higher energy savings on hardware. However, most of these works usually perform homogenous quantization or assign the same bit-width across all layers of a DNN that could be sub-optimal in terms of overall savings. In DNNs, each layer has a different structure that can lead to different properties related to quantization. Thus, some recent works have shown the possibility of assigning different bit-width to different layers of a DNN to achieve optimal quantization result \cite{chakraborty2020constructing}. In all the above works, quantization is carried out while ensuring that the overall accuracy of the network is maintained with the main focus of trading-off \textit{energy-and-accuracy} with some optimal compression methodology.

In this paper, we propose QUANOS- Adversarial Noise Sensitivity driven Hybrid Quantization for energy efficient, adversarially robust and accurate DNNs. We find the optimal bit-width of each DNN layer based on its sensitivity to adversarial perturbations. Conventionally, parameter importance of individual layers is assigned based on their impact on overall model accuracy. That is, parameters that impact the accuracy the most are represented using higher precisions (larger quantization widths), while low-impact parameters are represented with fewer bits. QUANOS, on the other hand, assigns parameter importance based on the layer’s contribution to adversarial perturbation. So, layers that contribute more to adversarial perturbations are quantized to lower precisions and vice-versa. Consequently, our method can reduce the model size significantly while maintaining a certain accuracy with the additional benefit of being adversarially robust. We propose a novel Adversarial Noise Sensitivity (ANS) metric to estimate the adversarial contribution of each layer. Using ANS, we avoid the exhaustive search for optimal bit-width per layer, and make the quantization process more efficient. Furthermore, QUANOS does not rely on having a pre-trained model to determine the bit compression. It is applied at the beginning of training (specifically, after $\sim$ 20-30  epochs). This allows us to train a model with reduced bit-width at each layer yielding an optimal quantized network with inference as well as training energy savings. 

To evaluate the benefits of QUANOS, we conduct a comprehensive hardware evaluation study. We design a 2-D systolic array accelerator to perform storage, computation analysis and calculate the energy efficiency and memory compression with QUANOS. Additionally, we evaluate the inference energy savings by integrating our accelerator with runtime configurable and precision scalable Multiply and Accumulate (MAC) architectures proposed in \cite{moons2016energy, moons201714}. Specifically, we compare between Dynamic Voltage-Accuracy-Frequency Scaling (DVAFS) \cite{moons201714}, and Data Gating (DG) \cite{moons2016energy}. This comparative analysis further highlights the relative benefits of QUANOS that performs hybrid layer specific quantization over homogeneously quantized networks. In summary, the key contributions of our work are as follows:
\begin{itemize}
\item{We propose QUANOS-  an accurate and efficient method to find optimal bit-width for each layer of a DNN that not only reduces the overall compute complexity, but also improves adversarial robustness.}
\item{We test QUANOS using benchmark datasets- CIFAR10, CIFAR100 and show that QUANOS achieves higher energy efficiency, adversarial robustness at near or iso-accuracy with respect to an 8-bit quantized baseline. At iso-compression rate, QUANOS yields higher adversarial robustness ($> 10\%-15\%$) than similar sized baseline. Evaluation of QUANOS on hardware platforms employing scalable MAC architectures yields $> 2\times$ improvement over 8-bit precision baseline.}
\end{itemize}

\section{QUANOS: Approach and Implementation}%Adversarial Noise Sensitivity: Definition and Outcome}
\subsection{Background on Adversarial Attacks}
\textbf{Generating Adversaries:} Adversarial examples are created using a trained DNN’s parameters and gradients. The adversarial perturbation, $\Delta$, is not just some random noise, but carefully designed to bias the network’s prediction on a given input towards a wrong class. Goodfellow et al. \cite{goodfellow2014explaining} proposed a simple method called Fast Gradient Sign Method (FGSM) to craft adversarial examples by linearizing a trained model’s loss function ($\mathcal{L}$, say cross-entropy loss) with respect to the input (X): 
\begin{equation}
X_{adv} = X + \epsilon \times sign(\nabla_X \mathcal{L}(\theta, X, y_{true})
\end{equation}

Here, $y_{true}$ is the true class label for the input $X,\theta$ denotes the model parameters (weights, biases etc.) and $\epsilon$ quantifies the magnitude of distortion. The net perturbation added to the input ($\Delta = \epsilon \times (\nabla_X \mathcal{L}(\theta, X, y_{true})$) is, thus, regulated by $\epsilon$. A key point to note here is that gradient propagation is a crucial step in adversarial input generation. This implies that deterring the flow of adversarial gradients can result in improved robustness. Further, the adversarial gradient contribution to the net perturbation ($\Delta$) from different layers can vary depending upon the learnt activations. This can enable us to evaluate the \textit{Adversarial Noise Sensitivity (ANS)} per layer. Besides FGSM, several other forms of attacks (essentially, multi-step variants of FGSM, like Projected Gradient Descent (PGD) \cite{madry2017towards} have also been proposed that cast stronger attacks. 

\textbf{Types of Attacks:} There are two kinds of attacks: BlackBox (BB), White-Box (WB) that are used to study adversarial robustness. WB adversaries are created using the target model’s parameters, that is, the attacker has full knowledge of a target model’s training information. BB attacks refer to the case when the attacker has no knowledge about the target model’s parameters.
%In this case, adversaries are created using a different source model’s parameters trained on the same classification task as the target model. Since BB attacks are transferred onto the target model, they are weaker than WB attacks. 
Security against WB attacks is a stronger notion and robustness against WB attacks guarantees robustness against BB for similar perturbation ($\epsilon$) range.  In all our experiments, we use WB attacks to verify the robustness of our proposed technique. Generally, a model’s adversarial robustness is quantified in terms of \textit{adversarial accuracy}. \textit{Adversarial accuracy} is the accuracy of a DNN on the adversarial dataset created using the test data for a given task. Higher accuracy implies more robustness. 

\subsection{Adversarial Noise Sensitivity (ANS)}
\begin{figure*}[h]
\centering
\includegraphics[width=\linewidth]{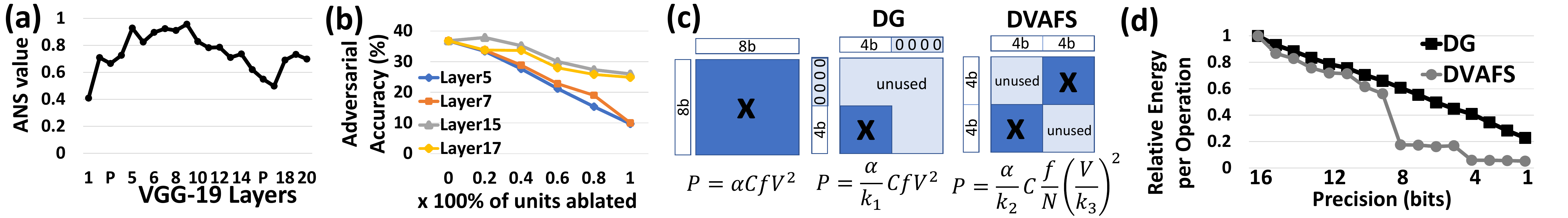}
\caption{(a) ANS value plotted across diferent layes of VGG-19 trained on CIFAR10. The x-axis denotes the layers as \{1,2,P,4,5...\}, where `P' denotes pooling layer. (b) Adversarial accuracy variation shown as different layers of VGG-19 model are ablated to different proportion. (c) Standard, DG, DVAFS configuration shown for the multiplication operation of a MAC accelerator processing 8-bit$\times$8-bit and scaled 4-bit$\times$4-bit operands. (d) Normalized energy/operation with precision scaling on DG, DVAFS architectures.}
\label{fig1}
\end{figure*}
A novel outcome of this paper is the identification of a new form of noise stability for DNNs, i.e., the sensitivity of each layer’s computation to adversarial noise. Fig. \ref{fig1} (a) visualizes the sensitivity of a VGG-19 model trained on CIFAR10 data. ANS is measured in terms of error ratio defined as:  
\begin{equation}
    ANS_l= \frac{||a_{adv}^l - a^l||_2}{||a^l||_2}
\end{equation}
Eqn. 2 measures the layerwise error ratio $ANS_l$. $a_{adv}^l,a^l$ are activation values of a layer $l$ when $X_{adv},X$ are presented to the network. Note, we show results for a DNN (composed of Rectified Linear Unit or ReLU neurons) with WB adversaries $X_{adv}$ created using FGSM. We see that the error ratio in Fig. \ref{fig1} (a) at the middle layers (\textit{layer 4-layer 10}) is significantly higher. In fact, \textit{layers 5,7} have the highest error measure. This error ratio is representative of ANS, wherein, higher ratio implies more changes in activations which can be attributed to higher adversarial gradient contribution by \textit{layers 5,7}. To further support our ANS results, we conducted an ablation study, that measures the importance of a single direction (or neuronal activation) to a network’s computation by asking how the network’s performance diminishes as the direction is removed. Specifically, we measured the network’s adversarial accuracy as we removed or clamped the activations of intermediate layers’ neurons, as shown in Fig. \ref{fig1} (b). We find that ablating random units in high ANS \textit{layers 5, 7} results in a drastic change in adversarial accuracy. On the other hand, ablating low ANS \textit{layers 15, 17} shows a very slow decline in accuracy. Even after zeroing out 99.99\% of the neurons in \textit{layers 15, 17}, the adversarial accuracy is pretty reasonable at $\sim 31\%, 28\%$, respectively. The steep decline in accuracy in the high ANS layer's ablation study implies that high ANS layers are more adversarially susceptible than low ANS layers. The consistency between Fig. \ref{fig1} (a), (b) illustrates ANS as a simple, yet, powerful metric to evaluate how does each layer contribute to the net adversarial perturbation during the gradient propagation. Based on this, we propose QUANOS where, we disrupt the flow of adversarial gradients by quantizing the individual DNN layers in proportion to their ANS values. High ANS layers are quantized to low precision and low ANS layers are maintained at high precision to yield robust DNNs.

\subsection{QUANOS: Hybrid Quantization based on ANS}
Despite their remarkable classification accuracies, large DNNs assimilate redundancies. Several recent works have studied these redundancies by abstracting the network from different levels and searching for, in particular, redundant filters (DenseNets \cite{huang2016densely}) and redundant connections (Deep Compression \cite{han2015deep}). In this work, we present a novel perspective on redundancy by tying it with adversarial robustness and use it for quantization of network parameters and activations. That is, we approximate the minimum size of the network when each parameter and activation of each individual layer is allowed to have a distinct number of precision bits. %Our goal here is to represent layers with high precisions only when they are critical to the adversarial accuracy of the network. %In this sense, our approach is similar to weight pruning (Han et al., 2015) which eliminates all but the essential parameters, producing a sparse network. 

Algorithm 1 outlines our technique. Initially we take a baseline $k_{l_{initial}}$-bit precision network and train it for a few epochs (generally, 30 epochs in all our experiments). Then, we calculate the $ANS_l$ value of each layer $l$ of the partially trained network. Based on the ANS values, we quantize each layer of the network as follows:
\begin{equation}
    k_l = k_{l_{initial}}- round(ANS_l *k_{l_{initial}}) 
\end{equation}
Say, ANS values of a 3-layered DNN are \{0.7, 0.4, 0.9\}. If the initial DNN has uniform 16-bit precision layers, applying Eqn. 3 will yield a quantized DNN with \{5-bit, 10-bit, 2-bit\} hybrid precision layers. Both weights and activations of a given layer are quantized to $k_l$ precision. Note, high ANS layers are quantized to lower precisions than low ANS layers. One major advantage of QUANOS is that it does not rely on a pre-trained model to perform optimal quantization. We can take a partially trained model and perform QUANOS analysis to obtain a compressed DNN which is then further trained till convergence (see Algorithm 1). Note, we can also apply QUANOS iteratively as: 1) Train a full precision DNN for a few epochs, 2) Compute $ANS$, 3) Quantize the DNN based on ANS, 4) Train the quantized DNN till convergence and then 5) Repeat Steps 2-4 till optimal accuracy-efficiency tradeoff is achieved. In all our experiments, we find that starting from a 16-bit baseline ($k_{l_{initial}} =16$) results in an optimal sized DNN with varying precisions ranging from 1-bit to 9-bit with just 1 iteration of QUANOS. Instead, using a 32-bit baseline ($k_{l_{initial}} =32$) as the starting point requires 2-3 iterations of QUANOS. Note, Algorithm 1 determines the precision of weights and activations of the convolutional layers of a network. The pooling layers following a convolutional layer automatically receive quantized activations. 
\begin{algorithm}
\SetAlgoLined
 Take a randomly initialized DNN (say, 16-bit) and train it for 20-30 epochs\;
 \For{each layer $l$ in DNN}{
   Compute $ANS_l$ using Eqn. 2\;
   $k_l = 16 - round(16*ANS_l)$ (Eqn. 3)\;
  }
  Train the $k_l$-bit hybrid precision DNN till convergence\;
 \caption{QUANOS Procedure}
\end{algorithm}

\section{Hardware Evaluation Setup}
%\begin{figure}[h]
%\centering
%\includegraphics[width=0.65\linewidth]{fig2_1.pdf}
%\caption{Simulation results for the network.}
%\label{fig2}
%\end{figure}
We evaluate the energy savings of QUANOS on a precision scalable MAC accelerator with different hardware scalability features such as DVAFS and DG. In an accelerator, the primary model dependent metrics that affect the energy consumption of a classification task are the MAC operations and memory accesses. Say, a particular convolutional layer of a DNN comprises of $I$ input channels, $O$ output channels, input map size $N \times N$ , weight kernel size $k \times k$ and output size $M \times M$. We illustrate the number of memory accesses and computations in a $k_b-bit$ layer in Table I. The energy consumption is calculated based on projections on 45nm CMOS technology for 32-bit precision operations. The energy consumed by any layer $l$ is given by
\begin{equation}
    E_l = N_{A-k_b}*E_{A-k_b}+N_{C-k_b}*E_{C-k_{b}}
\end{equation}
Note, we exclude the energy due to instruction flow and control flow. Further, the energy calculation in Eqn. 4 is a rather conservative estimate ignoring weight or input sharing. To exploit the benefits of hybrid precision, we designed MAC arithmetic circuits by integrating DG and DVAFS features. In DG, only MSB operands are used for computation while LSBs are kept at zero. This avoids unnecessary toggling in the circuit reducing switching activity as well as the critical path thereby enabling lower voltage operation for constant throughput. DVAFS logic reuses inactive cells at reduced precision scaling together both weight and activation with symmetric subword parallelism. Similar to DG, the shortened critical path permits lower supply voltage and lower frequency operation for constant throughput. Fig. \ref{fig1} (c) illustrates the DG and DVAFS multiplier operation for $4b \times 4b$ scaling of weights and activations in comparison to a $8b \times 8b$ baseline. Note the unused and used portions of the operands in each logic. In case of DG, the MAC block does not have any additional features such as selective clock gating. That is, all registers (MSB/LSB) stay clocked despite zeroing out of LSBs. DVAFS embeds additional design features for clock-gating and increased parallelism. 
\begin{table}[!t]
%% increase table row spacing, adjust to taste
%\renewcommand{\arraystretch}{1.3}
% if using array.sty, it might be a good idea to tweak the value of
% \extrarowheight as needed to properly center the text within the cells
\caption{Energy Calculation Chart for a given layer in a DNN}
\label{tab1}
\centering
%% Some packages, such as MDW tools, offer better commands for making tables
%% than the plain LaTeX2e tabular which is used here.
\begin{tabular}{|c|c|c|}
\hline
Operation & Term & Number of Operations\\
\hline
$k_b$ bit Memory access & $N_{A-k_b}$ & $N^2 \times I + k^2 \times I \times O$\\
$k_b$ bit MAC Computations& $N_{C-k_b}$ & $M^2 \times I \times k^2 \times O$\\
\hline
Operation & Term & Energy (pJ)\\
\hline
$k_b$ bit Memory access & $E_{A-k_b}$ & $2.5k_b$\\
32 bit MULT INT & $E_{M-I}$ & 3.1\\
32 bit ADD INT & $E_{Add-I}$ & 0.1\\
$k_b$ bit MAC INT & $E_{C-k_b}$ & $((3.1*k_b)/32 + 0.1)$\\
\hline
\end{tabular}
\end{table}

We synthesized the DVAFS-/DG-enabled MAC configuration circuits in 45nm process with a  nominal supply voltage of 1V from abstract-level System Verilog descriptions. Conservative power models were used for synthesis and power estimation. For DVAFS, we assessed the energy for each scaled mode of operation while sweeping the voltage from 1V to 0.8V. Fig. \ref{fig1} (d) shows the breakdown of energy per MAC operation for each configuration with respect to varying precision. Energy values are normalized with respect to a full 16-bit precision baseline MAC architecture. We combine the results of Fig. \ref{fig1} (d) and Eqn. 4 to estimate the total energy of a DNN layer when integrated with  DG/DVAFS hardware configuration as:
\begin{equation}
    E_{l_{config}} = N_{A-k_b}*E_{A-k_b}+N_{C-k_b}*E_{C-k_b}*E_{{config}_{k_b}}
\end{equation}
$E_{{config}_{k_b}}$ is the energy per operation evaluated for a given $k_b-bit$ precision and configuration as per Fig. \ref{fig1} (d).

Besides energy, we also evaluate the overall memory compression achieved with QUANOS. For a given layer convolutional $l$, the total memory required is equivalent to the product of total number of weights and the precision of the weights, given by: $M_l = I \times O \times k^2 \times k_b$. 

\section{Experimental Results}
We test QUANOS on two prevailing image classification benchmarks- CIFAR10 and CIFAR100. We use VGG-19 and ResNet18 models to conduct all analysis on CIFAR10 and CIFAR100, respectively. We imported github models and used similar hyperparameters and training methodologies as \cite{git} to conduct our experiments in PyTorch. For fair energy comparison, we use homogeneously quantized $8-bit$ and $16-bit$ precision models trained from scratch as baselines. We find that $8-bit$ and $16-bit$ models achieve iso-accuracy with that of a $32-bit$ model. We compare the energy/memory efficiency, adversarial accuracy (i.e. measure of robustness) and clean test accuracy of the final network obtained with QUANOS to that of the corresponding baselines. Note, as in most quantization works, we quantize the model's weights and activations to optimal bit-widths during forward propagation. The gradient calculation and update steps during backpropagation are done using full $32-bit$ precision. In QUANOS, as outlined in Algorithm 1, we train a $16-bit$ model initially for 30 epochs. Then, based on the ANS calculation, we quantize the individual layers of the DNN. In our experiments, we calculate the ANS metric based on adversarial inputs created using WB FGSM attack with $\epsilon =0.05$ for a random 1000-2000 sample of training inputs. 

\begin{figure}[!t]
\centering
\includegraphics[width=\linewidth]{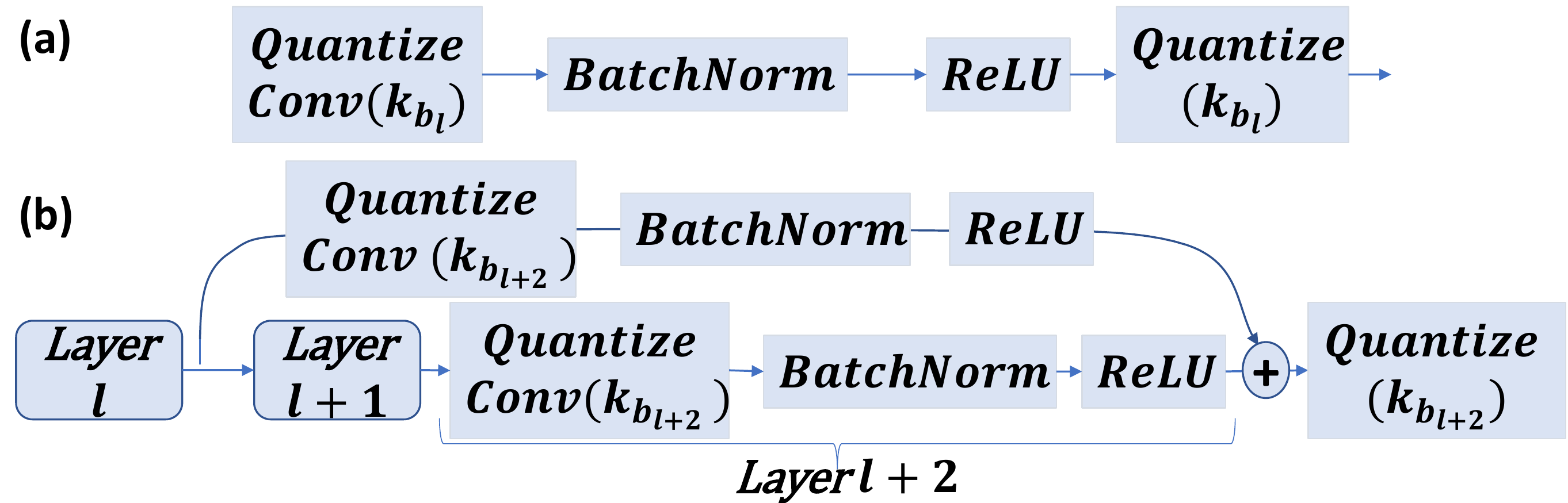}
\caption{Illustration of Quantization that is implemented with QUANOS on (a) standard block comprising convolutional, batch-norm, relu operation, (b) residual block with shortcut connection from layer $l$ to $l+2$.}
\label{fig3}
\end{figure}

\begin{figure*}[!h]
\centering
\includegraphics[width=0.85\linewidth]{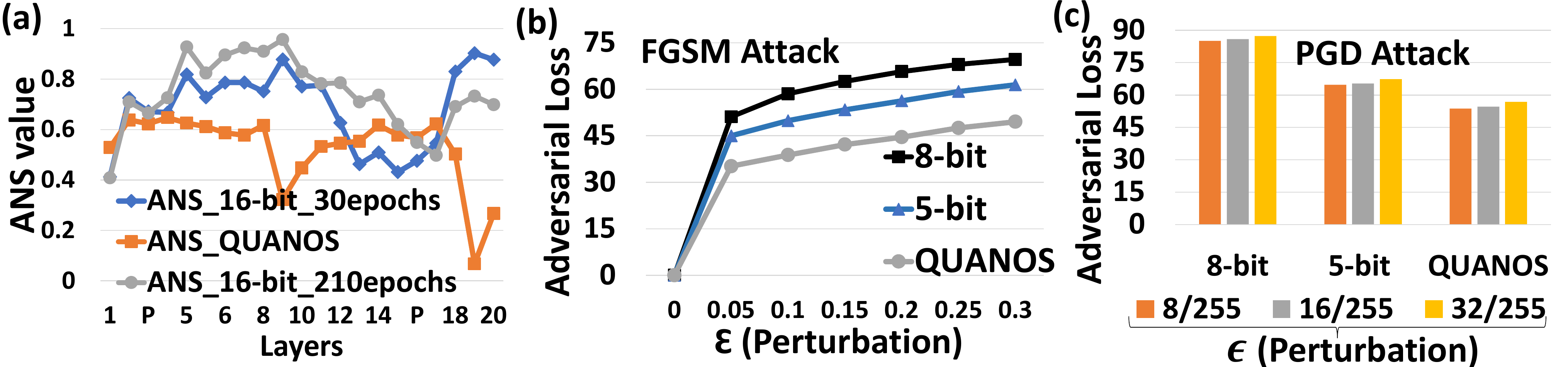}
\caption{(a) ANS values shown for QUANOS and 16-bit baseline VGG-19 model after training on CIFAR10 for certain number of epochs. Adversarial loss shown across different VGG-19 models trained on CIFAR10 for different perturbation strengths ($\epsilon$) against (b) FGSM attack, (c) PGD attack. }
\label{fig4}
\end{figure*}

\begin{table*}
%% increase table row spacing, adjust to taste
%\renewcommand{\arraystretch}{1.3}
% if using array.sty, it might be a good idea to tweak the value of
% \extrarowheight as needed to properly center the text within the cells
\caption{Result summary of CIFAR10 VGG-19 model.The bit-precision of each convolutional layer ($C1...C17$) is shown.}
\label{tab2}
\centering
%% Some packages, such as MDW tools, offer better commands for making tables
%% than the plain LaTeX2e tabular which is used here.
\resizebox{\textwidth}{!}{\begin{tabular}{|c|c|c|c|c|c|c|c|c|c|c|c|c|c|c|c|c|c|c|c|c|}
\hline

Model & \multicolumn{17}{|c|}{Layers} & Accuracy & Memory\\
 &C1 &C2 &C3& C4& C5 &C6 &C7&C8&C9&C10&C11&C12&C13&C14&C15&C16&C17 & & Compression\\
\hline
Baseline-1 & \multicolumn{17}{|c|}{-------------16-bit-------------} & 91.6&  0.32G (1$\times$)\\
\hline
Baseline-2 & \multicolumn{17}{|c|}{-------------8-bit-------------} & 91.4& 0.16G (0.5$\times$)\\
\hline
Baseline-3 & \multicolumn{17}{|c|}{-------------5-bit-------------} & 90.5& 0.1G (0.31$\times$) \\
\hline
QUANOS & 9-b & 4-b & 5-b & 3-b & 3-b& 3-b & 4-b & 2-b & 4-b & 6-b & 9-b &8-b &9-b &7-b &3-b &2-b &2-b  & 90.7& 0.09G (0.27$\times$)\\
\hline
\end{tabular}}
\end{table*}

Fig. 2 illustrates a quantized block of a DNN that is used for forward propagation in VGG, ResNet-like models. While quantization is pretty straightforward in VGG-like architectures, the residual connection in a ResNet model needs to be carefully quantized. Say, there is a shortcut connection between layer $l$ and layer $l+2$. We quantize the shortcut connection based on the ANS metric evaluated for layer $l+2$ i.e. the shortcut has the same precision as the layer into which it feeds. 

In our experiments, all layers of a DNN except the final output layer are quantized. The final output layer for a baseline/QUANOS model remains at $32-bit$ precision. We also train both baseline and QUANOS models for 210 epochs where accuracy saturation occurs. For robustness, we measure the adversarial accuracy of the models against WB FGSM and PGD attacks. PGD is one of the strongest attacks known in literature that casts adversaries over multiple steps \cite{madry2017towards}: $X_{adv}^{t+1} = \Pi(X_{adv}^{t} + \alpha (\nabla_X \mathcal{L}(\theta, X, y_{true})))$. In our experiments, we craft FGSM attacks for $\epsilon$ range- \{0.05, 0.1, 0.15, 0.2, 0.25, 0.3\}. For PGD, we craft attacks over $t=7$ steps of size $\alpha= 2/255$ over $\epsilon$ range-\{$8/255, 16/255, 32/255$\} as denoted in many recent works \cite{madry2017towards,panda2019discretization,lin2019defensive}.

\subsection{CIFAR10}
Table II illustrates the varying precision of different layers for a CIFAR10 VGG-19 model obtained with QUANOS. The average bit precision of the QUANOS model evaluated across all layers is 4.95 bits. For fair comparison of energy and adversarial accuracy at iso-compression, we trained a homogenously quantized 5-bit precision model for 210 epochs. The test accuracy of the baseline ($Baseline-1, Baseline-2, Baseline-3$ corresponding to 16-bit, 8-bit and 5-bit respectively) and the QUANOS model on clean inputs are comparable.  We plot the ANS values of the 16-bit baseline model (after 30 epochs of training as well as full training cycle of 210 epochs) and the model obtained with QUANOS (after full 210 epochs of training) in Fig. 3 (a). Note, the higher ANS layers are quantized to lower precisions. ANS quantifies the net adversarial perturbation contribution of different layers of a network. Interestingly, the overall ANS value for the QUANOS model is lower than that of the partially and fully trained baseline which implies that quantization curtails adversarial sensitivity.

\begin{table}
\caption{Energy results on CIFAR10 VGG-19 models of Table II for different hardware configurations.}
\label{tab2}
\centering
%% Some packages, such as MDW tools, offer better commands for making tables
%% than the plain LaTeX2e tabular which is used here.
\begin{tabular}{|c|c|c|c|}
\hline
Model & \multicolumn{3}{|c|}{Energy (in mJ)}\\
 & Standard & DG & DVAFS \\
\hline
Baseline-1 & 1.47 (1$\times$)& 1.47 (1$\times$)& 1.47 (1$\times$)\\
\hline
Baseline-2 & 0.75 (0.51$\times$)& 0.62 (0.42$\times$)& 0.47 (0.32$\times$)\\
\hline
Baseline-3& 0.49 (0.33$\times$) & 0.36 (0.24$\times$)& 0.29 (0.2$\times$)\\
\hline
QUANOS & 0.39 (0.26$\times$)&0.29 (0.2$\times$)& 0.25 (0.17$\times$)\\
\hline
\end{tabular}
\end{table}

\begin{table}
\caption{Comparison with adversarial training on CIFAR10 VGG-19 model of Table II. Adversarial loss shown for $\epsilon=\{2,8,16\}/255$ for FGSM, PGD attacks.}
\label{tab3}
\centering
%% Some packages, such as MDW tools, offer better commands for making tables
%% than the plain LaTeX2e tabular which is used here.
\begin{tabular}{|c|c|c|c|}
\hline
Scenario & Model& FGSM & PGD\\
\hline
\multirow{2}{*}{AdvTrain+FGSM} & 16-bit & 10.6, 39.6, 54 & 7.6, 49, 81\\
& QUANOS & 9, 31, 43 & 7, 44, 72 \\
\hline
\multirow{2}{*}{AdvTrain+PGD} & 16-bit & 12.6, 41, 56 & 11, 43, 67\\
& QUANOS & 8.5, 34.5, 52 & 6.5, 37.5, 65.5 \\
\hline
\end{tabular}
\end{table}

Fig. 3 (b), (c) compare the adversarial robustness of the QUANOS model with that of 16-bit $Baseline-1$ and 5-bit $Baseline-3$ against FGSM, PGD attacks. The plots in Fig. 3 (b), (v) show the overall \textit{Adversarial loss} measured as \textit{Clean Test Accuracy - Adversarial Accuracy}. Thus, lower loss implies higher robustness. We find that QUANOS consistently yields lower loss than the baseline models across varying $\epsilon$. We observe $\sim10\%$ ($\sim15\%$) higher robustness in FGSM (PGD) attacks with QUANOS than a similar sized 5-bit baseline.

\begin{table*}
%% increase table row spacing, adjust to taste
%\renewcommand{\arraystretch}{1.3}
% if using array.sty, it might be a good idea to tweak the value of
% \extrarowheight as needed to properly center the text within the cells
\caption{Result summary of CIFAR100 ResNet-18 model.The bit-precision of each convolutional layer ($C1...C17$) is shown.}
\label{tab2}
\centering
%% Some packages, such as MDW tools, offer better commands for making tables
%% than the plain LaTeX2e tabular which is used here.
\resizebox{\textwidth}{!}{\begin{tabular}{|c|c|c|c|c|c|c|c|c|c|c|c|c|c|c|c|c|c|c|c|c|}
\hline

Model & \multicolumn{17}{|c|}{Layers} & Accuracy & Memory\\
 &C1 &C2 &C3& C4& C5 &C6 &C7&C8&C9&C10&C11&C12&C13&C14&C15&C16&C17&  & Compression\\
\hline
Baseline-1 & \multicolumn{17}{|c|}{-------------16-bit-------------} & 76.7&  0.18G (1$\times$)\\
\hline
Baseline-2 & \multicolumn{17}{|c|}{-------------8-bit-------------} & 74& 0.09G (0.5$\times$)\\
\hline
Baseline-3 & \multicolumn{17}{|c|}{-------------4-bit-------------} & 72.1& 0.05G (0.25$\times$) \\
\hline
QUANOS & 9-b & 8-b & 7-b & 4-b & 5-b& 3-b & 4-b & 2-b & 3-b & 2-b & 3-b &2-b &2-b &2-b &2-b &2-b &2-b & 72.8 & 0.024G (0.14$\times$)\\
\hline
\end{tabular}}
\end{table*}

\begin{table}
\caption{Energy results on CIFAR100 ResNet18 models of Table V for different hardware configurations.}
\label{tab2}
\centering
\begin{tabular}{|c|c|c|c|}
\hline

Model & \multicolumn{3}{|c|}{Energy (in mJ)}\\
 & Standard & DG & DVAFS \\
\hline
Baseline-1 & 1.44 (1$\times$)& 1.44 (1$\times$)& 1.44 (1$\times$)\\
\hline
Baseline-2 & 0.75 (0.52$\times$)& 0.45 (0.32$\times$)& 0.13 (0.09$\times$)\\
\hline
Baseline-3& 0.40 (0.28$\times$) & 0.165 (0.114$\times$)& 0.023 (0.02$\times$)\\
\hline
QUANOS & 0.32 (0.22$\times$)&0.17 (0.118$\times$)& 0.095 (0.07$\times$)\\
\hline
\end{tabular}
\end{table}

The net energy calculated using Eqn. 4, 5 for different configurations are also shown in Table III. \textit{Standard} corresponds to the accelerator configuration without any hardware scalability features. The MAC hardware architecture is designed for a 16-bit baseline. Hence, integrating DG, DVAFS does not change the overall energy consumption of the 16-bit precision model. We start observing savings by lowering the bit-precision of the model (Table III). QUANOS yields $\sim 3.8\times-5.9\times$ energy reduction over the 16-bit baseline and $\sim 1.8\times-2\times$ lower energy than 8-bit baseline at iso-accuracy. Table II also shows the memory compression results. QUANOS yields $\sim 3.7\times$ ($1.2\times$) higher compression than 8-bit (5-bit) baseline, respectively. The advantage of memory compression with QUANOS is that the training cost incurred in the hybrid model will also be $\sim 3.7\times$ lower than the 8-bit model baseline at iso-epochs.

In Fig. \ref{fig4} (b), (c), we see that adversarial attacks cause sharp increase in adversarial loss. While QUANOS provides an inherent resilience, strong attacks with high perturbation strengths ($\epsilon$) have devastating effects. To mitigate the adversarial effects, we can integrate our model with defense mechanisms. Adversarial training \cite{madry2017towards} is currently the strongest method for defense. By augmenting the training set with adversarial samples, the network learns to classify adversarial samples correctly. Augmenting the dataset with adversaries created using WB FGSM attacks is referred to as \textit{AdvTrain+FGSM}. Similarly, augmenting dataset with WB PGD adversaries is referred as \textit{AdvTrain+PGD}. It is evident AdvTrain+PGD will provide a stronger defense than AdvTrain+FGSM since we train the network with stronger adversaries for the former.

Table IV compares the adversarial loss of a 16-bit baseline and QUANOS model (shown in Table II) trained with adversarial data augmentation. In case of QUANOS, we perform adversarial training after conducting ANS analysis and hybridizing the layers with varied precision. While adversarial training substantially improves the robustness of both models, QUANOS yields $\sim1\%-10\%$ and $\sim4\%$ lower adversarial loss than the 16-bit baseline against FGSM and PGD attacks across different $\epsilon$, respectively. Thus, integrating QUANOS with state-of-the-art defense techniques can amplify robustness even outperforming the state-of-the-art methods.

\subsection{CIFAR100}
\begin{figure}[h]
\centering
\includegraphics[width=\linewidth]{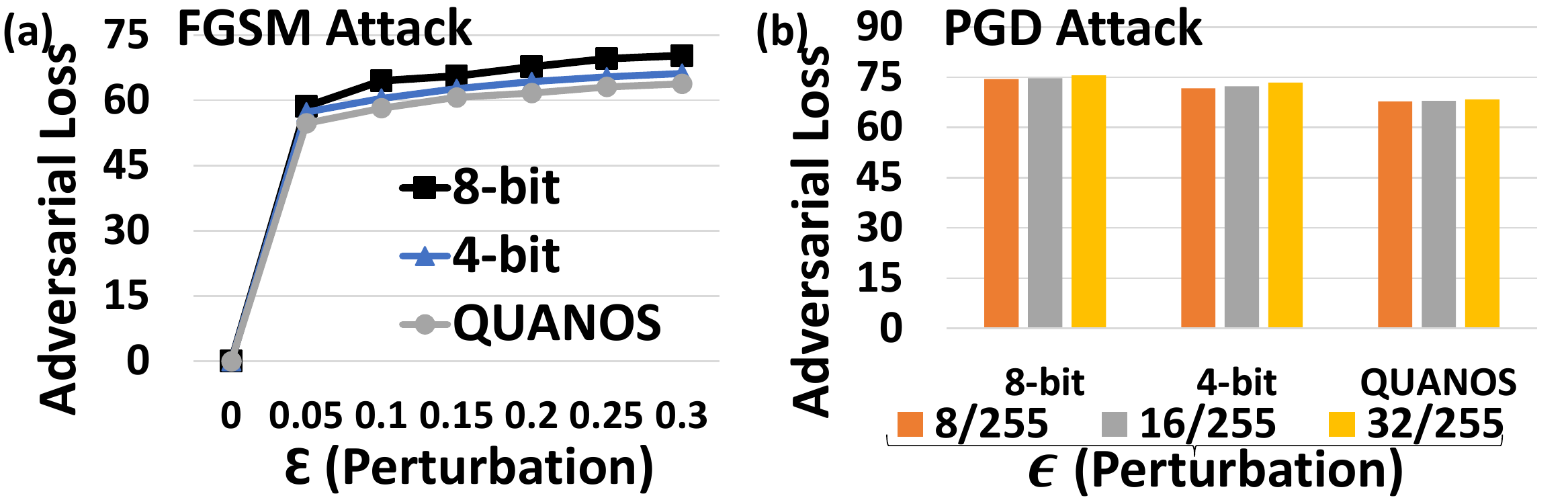}
\caption{Adversarial loss shown across different ResNet18 models trained on CIFAR100 for different $\epsilon$ against (b) FGSM attack, (c) PGD attack.}
\label{fig3}
\end{figure}

Table V, VI show the accuracy, memory compression and overall energy results for CIFAR100 ResNet18 models across different bit-precision scenarios: \textit{Baseline-1 (16-bit), Baseline-2 (8-bit), Baseline-3 (4-bit)} and \textit{QUANOS (Hybrid)}. The bit-precision of the different layers of the QUANOS model averages to $\sim3.65$. Thus, we perform iso-compression comparison between QUANOS and Baseline-3. Overall, QUANOS yields higher energy efficiency and memory compression benefits at near iso-accuracy than all baseline models in most cases ($3.6\times$ memory compression, $1.4\times-2.4\times$ energy efficiency than 8-bit Baseline-2). We find that for \textit{DG, DVAFS} hardware configuration, QUANOS yields slightly higher energy than Baseline-3 (Table VI). The higher energy in QUANOS is a consequence of its high precision ($>4-bit$) initial layers $C1 - C5$ (Table V). Note, as explained in Fig. 2, the residual connections in QUANOS are scaled to the same bit-precison as the layer it feeds into. That is, according to Table V, a shortcut connection between C2 (8-bit) and C4 (4-bit) will be quantized to 4-bit.

Fig. 4 (a), (b) show the adversarial loss of the models across FGSM, PGD attacks. QUANOS yields $4\%-7\%$ lower loss than 8-bit baseline. QUANOS and 4-bit model exhibit similar robustness to FGSM attacks, while, QUANOS is more resilient against PGD attacks. Table VII shows how QUANOS models trained with adversarial training perform with respect to a adversarially trained 4-bit model (Baseline-3) against strong PGD attacks. QUANOS has significantly ($>10\%$ adversarial loss difference for large $\epsilon$ PGD attacks) higher adversarial robustness than 4-bit Baseline-3.

\begin{table}
%% increase table row spacing, adjust to taste
%\renewcommand{\arraystretch}{1.3}
% if using array.sty, it might be a good idea to tweak the value of
% \extrarowheight as needed to properly center the text within the cells
\caption{Comparison with adversarial training on CIFAR100 ResNet18 model from Table V. Adversarial loss shown for $\epsilon=\{2,8,16\}/255$ for FGSM, PGD attacks.}
\label{tab3}
\centering
\begin{tabular}{|c|c|c|c|}
\hline
Scenario & Model& FGSM & PGD\\
\hline
\multirow{2}{*}{AdvTrain+FGSM} & 4-bit & 12.6, 43.6, 58.6 & 9.6, 66, 90\\
& QUANOS & 8.6, 37.6, 56.6 & 5.6, 19, 34 \\
\hline
\multirow{2}{*}{AdvTrain+PGD} & 4-bit & 13, 21.3, 25.8 & 15.6, 26.6, 30.3\\
& QUANOS & 11.2, 16.3, 20 & 9, 10.4, 14.9 \\
\hline
\end{tabular}
\end{table}

\section{Conclusion}
We present QUANOS- a structured method for hybrid quantization of different layers of a deep neural network to produce energy-efficient, accurate and adversarially robust models. We propose a novel metric, adversarial noise sensitivity (ANS), that evaluates the contribution of each layer towards adversarial noise and then, determines the optimal bit-width per layer. We perform energy evaluation of quantized models on hardware architectures integrated with data gating (DG) and dynamic voltage accuracy frequency scaling (DVAFS) features. Our experiments on CIFAR10, CIFAR100 datasets reveal QUANOS yields significant benefits than homogeneously quantized 8-bit and similar sized baseline models.

%references section

%\bibliographystyle{IEEEtran}
%\bibliography{IEEEabrv,bib}


\end{document}